\documentclass[10pt,journal,compsoc]{IEEEtran}
\ifCLASSOPTIONcompsoc
  \usepackage[nocompress]{cite}
\else
  \usepackage{cite}
\fi

\ifCLASSINFOpdf
   \usepackage[pdftex]{graphicx}
   \graphicspath{{./images/}}
   \DeclareGraphicsExtensions{.pdf,.jpeg,.png}
\else
   \usepackage[dvips]{graphicx}
   \graphicspath{{./images/}}
   \DeclareGraphicsExtensions{.eps}
\fi

\usepackage{amsmath}
\usepackage{svg}
\usepackage{amsfonts}
\usepackage{algorithm}
\usepackage[noend]{algpseudocode}
\usepackage{subfig}

\begin{document}
\title{EEG-Based Emotion Recognition Using Regularized Graph Neural Networks}

\author{Peixiang~Zhong,
        Di~Wang,~\IEEEmembership{Senior Member,~IEEE,}
        and~Chunyan~Miao,~\IEEEmembership{Senior Member,~IEEE}
\IEEEcompsocitemizethanks{\IEEEcompsocthanksitem P. Zhong, D. Wang and C. Miao are with the Joint NTU-UBC Research Centre of Excellence in Active Living for the Elderly (LILY), Nanyang Technological University, Singapore.\protect\\
P. Zhong and C. Miao are also with the Alibaba-NTU Singapore Joint Research Institute and the School of Computer Science and Engineering, Nanyang Technological University, Singapore.\protect\\
E-mail: peixiang001@e.ntu.edu.sg, \{wangdi, ascymiao\}@ntu.edu.sg
}
}


\IEEEtitleabstractindextext{%
\begin{abstract}
Electroencephalography (EEG) measures the neuronal activities in different brain regions via electrodes. Many existing studies on EEG-based emotion recognition do not fully exploit the topology of EEG channels. In this paper, we propose a regularized graph neural network (RGNN) for EEG-based emotion recognition. RGNN considers the biological topology among different brain regions to capture both local and global relations among different EEG channels. Specifically, we model the inter-channel relations in EEG signals via an adjacency matrix in a graph neural network where the connection and sparseness of the adjacency matrix are inspired by neuroscience theories of human brain organization. In addition, we propose two regularizers, namely node-wise domain adversarial training (NodeDAT) and emotion-aware distribution learning (EmotionDL), to better handle cross-subject EEG variations and noisy labels, respectively. Extensive experiments on two public datasets, SEED and SEED-IV, demonstrate the superior performance of our model than state-of-the-art models in most experimental settings. Moreover, ablation studies show that the proposed adjacency matrix and two regularizers contribute consistent and significant gain to the performance of our RGNN model. Finally, investigations on the neuronal activities reveal important brain regions and inter-channel relations for EEG-based emotion recognition. 
\end{abstract}

}

\maketitle

\IEEEdisplaynontitleabstractindextext
\IEEEpeerreviewmaketitle

\IEEEraisesectionheading{\section{Introduction}\label{sec: introduction}}

\IEEEPARstart{E}{motion} recognition focuses on the recognition of human emotions based on a variety of modalities, such as audio-visual expressions, body language, physiological signals, etc. Compared to other modalities, physiological signals, such as  electroencephalography (EEG),  electrocardiogram (ECG), electromyography (EMG), etc., have the advantage of being difficult to hide or disguise. In recent years, due to the rapid development of noninvasive, easy-to-use and inexpensive EEG recording devices, EEG-based emotion recognition has received an increasing amount of attention in both research \cite{alarcao2017emotions} and applications \cite{acharya2015computer}.

Emotion models can be broadly categorized into discrete models and dimensional models. The former categorizes emotions into discrete entities, e.g., anger, disgust, fear, happiness, sadness, and surprise in Ekman's theory \cite{ekman1997universal}. The latter describes emotions using their underlying dimensions, e.g., valence, arousal and dominance \cite{mehrabian1996pleasure}, which measures emotions from unpleasant to pleasant, passive to active, and submissive to dominant, respectively.

EEG signals measure voltage fluctuations from the cortex in the brain and have been shown to reveal important information about human emotional states \cite{schmidt2001frontal}. For example, greater relative left frontal EEG activity has been observed when experiencing positive emotions \cite{schmidt2001frontal}. The voltage fluctuations in different brain regions are measured by electrodes attached to the scalp. Each electrode collects EEG signals in one channel. The collected EEG signals are often analyzed in specific frequency bands, namely delta (1-4 Hz), theta (4-7 Hz), alpha (8-13 Hz), beta (13-30 Hz), and gamma (\textgreater 30 Hz).

Many existing EEG-based emotion recognition methods are primarily based on the supervised machine learning approach, wherein features are often extracted from preprocessed EEG signals in each channel over a time window. Then, a classifier is trained on the extracted features to recognize emotions. Wang \textit{et al.} \cite{wang2014emotional} compared power spectral density features (PSD), wavelet features and nonlinear dynamical features with a Support Vector Machine (SVM) classifier. Zheng and Lu \cite{zheng2015investigating} investigated critical frequency bands and channels using PSD, differential entropy (DE) \cite{shi2013differential} and PSD asymmetry features, and obtained robust accuracy using deep belief networks (DBN). However, most existing EEG-based emotion recognition approaches do not address the following three challenges: 1)~the topological structure of EEG channels are not effectively exploited to learn more discriminative EEG representations; 2)~EEG signals vary significantly across different subjects, which hinders the generalizability of the trained classifiers in subject-independent classification settings; and 3)~participants may not always generate the intended emotions when watching emotion-eliciting stimuli. Consequently, the emotion labels in the collected EEG data may be noisy and inconsistent with the actual elicited emotions. 

There have been several attempts to address the first challenge. Zhang \textit{et al.} \cite{zhang2018cascade} and Zhang \textit{et al.} \cite{zhang2018spatial} incorporated spatial relations in EEG signals using convolutional neural networks (CNN) and recurrent neural networks (RNN), respectively. However, their approaches require a 2D representation of EEG channels on the scalp, which may cause information loss during flattening because channels are actually arranged in the 3D space. In addition, their approach of using CNNs and RNNs to capture inter-channel relations has difficulty in learning long-range dependencies \cite{pascanu2013difficulty}. Graph neural networks (GNN) has been applied in \cite{song2018eeg} to capture inter-channel relations using an adjacency matrix. However, similar to CNNs and RNNs, the GNN approach \cite{song2018eeg} only considers relations between the nearest channels, which thus may lose valuable information between distant channels, such as the PSD asymmetry between channels on the left and right hemispheres in the frontal region, which has been shown to be informative in valence prediction \cite{schmidt2001frontal}. 

In recent years, several studies \cite{zheng2016personalizing, chai2017fast} attempted to tackle the second challenge by investigating the transferability of EEG-based emotion recognition models across subjects. Lan \textit{et al.} \cite{lan2018domain} compared several domain adaptation techniques such as maximum independence domain adaptation (MIDA), transfer component analysis (TCA), subspace alignment (SA), etc. They found that the subject-independent classification accuracy can be improved by around 10\%. Li \textit{et al.} \cite{li2018bi} applied domain adversarial training to lower the influence of individual subject on EEG data and obtained improved performance as well. However, their adversarial training does not exploit any graph structure of the EEG signals and only leads to small performance improvement in our experiment (see Section \ref{sec: ablation study}). 

To the best of our knowledge, no attempt has been made to address the third challenge, i.e., noisy emotion labels, in EEG-based emotion recognition.

In this paper, we propose a regularized graph neural network (RGNN) aiming to address all the three aforementioned challenges. Graph analysis for human brain has been studied extensively in the neuroscience literature \cite{bullmore2009complex, fornito2013graph}. However, making an accurate  connectome is still an open question and subject to different scales \cite{fornito2013graph}. Inspired by \cite{bullmore2012economy, song2018eeg}, we consider each EEG channel as a node in our graph. Our RGNN model extends the simple graph convolution network (SGC) \cite{wu2019simplifying} and leverages the topological structure of EEG channels. Specifically, we propose a sparse adjacency matrix to capture both local and global inter-channel relations based on the biological topology of human brain \cite{bullmore2012economy}. Local inter-channel relations connect nearby groups of neurons and may reveal anatomical connectivity at macroscale \cite{craddock2013imaging, fornito2013graph}. Global inter-channel relations connect distant groups of neurons between the left and right hemispheres and may reveal emotion-related functional connectivity \cite{schmidt2001frontal, li2018bi}. 

In addition, we propose a node-wise domain adversarial training (NodeDAT) method to regularize RGNN for better generalization in subject-independent classification scenarios. Different from the domain adversarial training in \cite{ganin2016domain, li2018bi}, our NodeDAT method provides a finer-grained regularization by minimizing the domain discrepancies between features in the source and target domains for each channel/node. Moreover, we propose an emotion-aware distribution learning (EmotionDL) method to address the problem of noisy labels in the datasets. Prior studies have shown that noisy labels can adversely impact classification accuracy \cite{zhu2004class}. Instead of learning the traditional single-label classification, our EmotionDL method learns a distribution of labels of the training data and thus acts as a regularizer to improve the robustness of our model against noisy labels. Finally, we conduct extensive experiments to validate the effectiveness of our RGNN model and investigate emotion-related informative neuronal activities.

In summary, the main contributions of this paper are as follows:
\begin{enumerate}
    \item We propose a regularized graph neural network (RGNN) model to recognize emotions based on EEG signals. Our biologically inspired model captures both local and global inter-channel relations.
    \item We propose two regularizers: node-wise domain adversarial training (NodeDAT) and emotion-aware distribution learning (EmotionDL), to improve the robustness of our model against cross-subject variations and noisy labels, respectively.
    \item We conduct extensive experiments in both subject-dependent and subject-independent classification settings on two public EEG datasets, namely SEED \cite{zheng2015investigating} and SEED-IV \cite{zheng2018emotionmeter}. Experimental results demonstrate the effectiveness of our proposed model and regularizers. In addition, our RGNN model achieves superior performance over the state-of-the-art models in most experimental settings. 
    \item We investigate the emotional neuronal activities and the results reveal that pre-frontal, parietal and occipital regions may be the most informative regions for emotion recognition. In addition, global inter-channel relations between the left and right hemispheres are important, and local inter-channel relations between (FP1, AF3), (F6, F8) and (FP2, AF4) may also provide useful information.
\end{enumerate}

\section{Related Work}
\label{sec: related work}
In this section, we review related work in the fields of EEG-based emotion recognition, graph neural network, unsupervised domain adaptation, and learning with noisy labels.
\subsection{EEG-Based Emotion Recognition}
EEG feature extractors and classifiers are the two fundamental components in the machine learning approach of EEG-based emotion recognition. EEG features can be broadly divided into single-channel features and multi-channel ones \cite{jenke2014feature}. The majority of existing features are computed on a single channel, e.g., statistical features \cite{tang2017eeg}, PSD \cite{lin2010eeg}, differential entropy (DE) \cite{shi2013differential}, and wavelet features \cite{akin2002comparison}. A few number of features are computed on multiple channels to capture the inter-channel relations, e.g., the asymmetry of PSD between two hemispheres \cite{zheng2015investigating} and functional connectivity \cite{wu2019identifying, li2019eeg}, where common indices such as correlation, coherence and phase synchronization were used estimate brain functional connectivity between channels. Another line of research in multi-channel features is to use common spatial filters \cite{wu2014probabilistic} and spatial-temporal filters \cite{wu2015bayesian, qi2015rstfc} to extract class-discriminative EEG features. In contrast, our model is deigned to operate on single-channel features and learn to effectively combine them using a graph neural network.

EEG classifiers can be broadly divided into topology-invariant classifiers and topology-aware ones. The majority of existing classifiers are topology-invariant classifiers such as SVM,  k-Nearest Neighbors (KNN), DBN \cite{zheng2014eeg} and RNN \cite{kim2018deep}, which do not take the topological structure of EEG features into account when learning the EEG representations. In contrast, topology-aware classifiers such as CNN \cite{li2016emotion, zhang2018cascade, li2018hierarchical} and GNN \cite{song2018eeg} consider the inter-channel topological relations and learn EEG representations for each channel by aggregating features from nearby channels using convolutional operations either in the Euclidean space or in the non-Euclidean space. However, as discussed in Section~\ref{sec: introduction}, existing CNNs and GNNs have difficulty in learning the dependencies between distant channels, which may reveal important emotion-related information. Recently, Zhang \textit{et al.} \cite{zhang2018spatial} and Li \textit{et al.} \cite{li2019novel} proposed to use RNNs to learn spatial topological relations between channels by scanning electrodes in both vertical and horizontal directions. However, their approaches do not fully exploit the topological structure of EEG channels. For example, two topologically close channels may be far away from each other in their scanning sequence. In contrast, our model is able to learn relations between distant channels using global connections.

\subsection{Graph Neural Network}
Graph neural network (GNN) is a type of neural network dealing with data in the graph domain, e.g., molecular structures, social networks and knowledge graphs \cite{wu2019comprehensive}. One early work on GNN \cite{scarselli2008graph} aimed to learn a converged static state embedding for each node in the graph using a transition function applied to its neighborhood. Later, inspired by the convolutional operation of CNN in the Euclidean domain, Bruna \textit{et al.} \cite{bruna2013spectral} combined spectral graph theory \cite{chung1997spectral} with neural network and defined convolutional operations in the graph domain using the spectral filters computed from the normalized graph Laplacian. Following this line of research, Defferrard \textit{et al.} \cite{defferrard2016convolutional} proposed fast localized convolutions by using a recursive formulation of the $K$-order Chebyshev polynomials to approximate the filters. The resulting representation for each node is an aggregation of its $K^{\text{th}}$-order neighborhood. Kipf and Welling \cite{kipf2017semi} further limited $K=1$ and proposed the standard graph convolutional network (GCN) with a faster localized graph convolutional operation. The convolutional layers in GCN can be stacked $K$ times to effectively convolve the $K^{\text{th}}$-order neighborhood of a node. Recently, Wu \textit{et al.} \cite{wu2019simplifying} simplified GCN by removing the nonlinearities between convolutional layers in GCN and proposed the simple graph convolution network (SGC), which effectively behaves like a linear feature transformation followed by a logistic regression. Apart from the convolution operation used in GCNs, there are other types of operations used in GNNs, such as attention \cite{veličković2018graph}. However, they are often trained significantly slower than SGC \cite{wu2019simplifying}. In this paper, we extend SGC to model EEG signals because it performs orders of magnitude faster than other networks with comparable classification accuracy.

\subsection{Unsupervised Domain Adaptation}
Unsupervised domain adaptation aims to mitigate the domain shift in knowledge transfer from a supervised source domain to an unsupervised target domain. The most common approaches are instance re-weighting and domain-invariant feature learning. Instance re-weighting methods \cite{huang2007correcting} aim to infer the resampling weight directly by feature distribution matching across source and target domains in a non-parametric manner. Domain-invariant feature leaning methods align features from both source and target domains to a common feature space. The alignment can be achieved by minimizing divergence \cite{gretton2012kernel}, maximizing reconstruction \cite{ghifary2016deep}, or adversarial training \cite{ganin2016domain}. Our proposed NodeDAT regularizer extends the domain adversarial training \cite{ganin2016domain} to graph neural networks and achieves finer-grained regularization by minimizing the discrepancies between features in source and target domains for each node individually.

\subsection{Learning with Noisy Labels}
Commonly adopted approaches to learning with noisy labels are based on the noise transition matrix and robust loss functions. The noise transition matrix specifies the probabilities of transition from each ground truth label to each noisy label and is often applied to modify the cross-entropy loss. The matrix can be pre-computed as \textit{a prior} \cite{patrini2017making} or estimated from noisy data \cite{sukhbaatar2014training}. A few studies tackle noisy labels by using noise-tolerant robust loss functions, such as unhinged loss \cite{van2015learning} and ramp loss \cite{brooks2011support}. 
Our proposed EmotionDL regularizer is inspired by \cite{gao2017deep}, which applies distribution learning to classify ambiguous images.

\section{Preliminaries}
\label{sec: preliminary}
In this section, we introduce the preliminaries of the simple graph convolution network (SGC) \cite{wu2019simplifying} and spectral graph convolution, which are the basis of our RGNN model.

\subsection{Simple Graph Convolution Network (SGC)}
\label{sec: sgc}
Given a graph $\mathcal{G}=(\mathcal{V}, \mathcal{E})$, where $\mathcal{V}$ denotes a set of nodes and $\mathcal{E}$ denotes a set of edges between nodes in $\mathcal{V}$. Data on $\mathcal{V}$ can be represented by a feature matrix $\mathbf{X} \in \mathbb{R}^{n\times d}$, where $n=|\mathcal{V}|$ and $d$ denotes the input feature dimension. The edge set $\mathcal{E}$ can be represented by a weighted adjacency matrix $\mathbf{A} \in \mathbb{R}^{n\times n}$ with self-loops, i.e., $\mathbf{A}_{ii} = 1$, $i=1,2,..., n$. In general, GNNs learn a feature transformation function for $\mathbf{X}$ and produces output $\mathbf{Z} \in \mathbb{R}^{n\times d^{'}}$, where $d^{'}$ denotes the output feature dimension.

Between adjacent layers in GNNs, the feature transformation can be written as 
\begin{equation}
    \mathbf{H}^{l+1} = f(\mathbf{H}^l, \mathbf{A}),
\end{equation}
where $l=0, 1, ..., L-1$, $L$ denotes the number of layers, $\mathbf{H}^0 = \mathbf{X}$, $\mathbf{H}^L = \mathbf{Z}$, and $f$ denotes the function we want to learn. A simple definition of $f$ would be 
\begin{equation}
\label{eqn: simple f}
    \mathbf{H}^{l+1} = \sigma (\mathbf{A} \mathbf{H}^l \mathbf{W}^l),
\end{equation}
where $\sigma$ denotes a non-linear function and $\mathbf{W}^l$ denotes a weight matrix at layer $l$. For each node $\mathbf{x}$, function $f$ simply computes the weighted sum of all the node features in its neighborhood including $\mathbf{x}$ itself, followed by a non-linear transformation. However, one major limitation of the $f$ in (\ref{eqn: simple f}) is that repeatedly applying $f$ along multiple layers may lead to $\mathbf{H}^l$ with overly large values due to summation. Kipf and Welling \cite{kipf2017semi} alleviated this limitation by proposing the graph convolution network (GCN) as follows
\begin{equation}
\label{eqn: gcn}
    \mathbf{H}^{l+1} = \sigma (\mathbf{D}^{-\frac{1}{2}} \mathbf{A} \mathbf{D}^{-\frac{1}{2}} \mathbf{H}^l \mathbf{W}^l),
\end{equation}
where $\mathbf{D}$ denotes the diagonal degree matrix of $\mathbf{A}$, i.e., $\mathbf{D}_{ii} = \sum_j{\mathbf{A}_{ij}}$. The normalized adjacency matrix $\mathbf{D}^{-\frac{1}{2}} \mathbf{A} \mathbf{D}^{-\frac{1}{2}}$ prevents $\mathbf{H}$ from growing overly large. If we ignore $\sigma$ and $\mathbf{W}^l$ temporarily and expand (\ref{eqn: gcn}), the hidden state $\mathbf{H}_i^{l+1}$ for node $\mathbf{x}_i$, $i=1,2, ..., n$, can be computed via
\begin{equation}
\label{eqn: element-wise feature propagation}
    \mathbf{H}_i^{l+1} \leftarrow \frac{\mathbf{A}_{ii}}{\mathbf{D}_{ii} + 1} \mathbf{H}_i^{l} + \sum_{j=1}^{n}\frac{\mathbf{A}_{ij}}{\sqrt{(\mathbf{D}_{ii} + 1) (\mathbf{D}_{jj} + 1)}} \mathbf{H_j}^{l}.
\end{equation}
Note that each neighboring $\mathbf{H}_j^l$ is now normalized by the degrees of both $\mathbf{x}_i$ and $\mathbf{x}_j$. 
Successively applying $L$ layers aggregates node features within a neighborhood of size $L$. 

To further accelerate training while keeping comparable performance, Wu \textit{et al.} \cite{wu2019simplifying} proposed SGC by removing the non-linear function $\sigma$ in (\ref{eqn: gcn}) and reparameterizing all linear transformations $\mathbf{W}^l$ across all layers into one linear transformation $\mathbf{W}$ as follows
\begin{equation}
\label{eqn: sgc}
\begin{split}
    \mathbf{Z} =\mathbf{H}^{L}= \mathbf{S} \mathbf{H}^{L-1} \mathbf{W}^{L-1}=...=\mathbf{S}^L \mathbf{X} \mathbf{W},
\end{split}
\end{equation}
where $\mathbf{S} = \mathbf{D}^{-\frac{1}{2}} \mathbf{A} \mathbf{D}^{-\frac{1}{2}}$ and $\mathbf{W}=\mathbf{W}^{0}\mathbf{W}^{1}...\mathbf{W}^{L-1}$. Essentially, SGC computes a topology-aware linear transformation $\mathbf{\hat{X}} = \mathbf{S}^L \mathbf{X}$, followed by a final linear transformation $\mathbf{Z} = \mathbf{\hat{X}}\mathbf{W}$.

\subsection{Spectral Graph Convolution}
\label{sec: spectral analysis}
We analyze GCN from the perspective of spectral graph theory \cite{chung1997spectral}. Graph Fourier analysis relies on the graph Laplacian $\mathbf{L} = \mathbf{D}-\mathbf{A}$ or the normalized graph Laplacian $\mathbf{\hat{L}} = \mathbf{I} - \mathbf{D}^{-\frac{1}{2}} \mathbf{A} \mathbf{D}^{-\frac{1}{2}}$. Since $\mathbf{\hat{L}}$ is a symmetric positive semidefinite matrix, it can be decomposed as $\mathbf{\hat{L}}=\mathbf{U}\mathbf{\Lambda} \mathbf{U}^T$, where $\mathbf{U}$ is the orthonormal eigenvector matrix of $\mathbf{\hat{L}}$ and $\mathbf{\Lambda} = \text{diag}(\lambda_1, ..., \lambda_N)$ is the diagonal matrix of corresponding eigenvalues. Given graph data $\mathbf{X}$, the graph Fourier transform of $\mathbf{X}$ is $\mathbf{\hat{X}}=\mathbf{U}^T \mathbf{X}$, and the inverse Fourier transform of $\mathbf{\hat{X}}$ is $\mathbf{X}=\mathbf{U}\mathbf{\hat{X}}$. Hence, the graph convolution between $\mathbf{X}$ and a filter $\mathbf{G}$ is computed as follows
\begin{equation}
\label{eqn: graph convolution}
    \mathbf{X} * \mathbf{G} = \mathbf{U}((\mathbf{U}^T \mathbf{G})\odot(\mathbf{U}^T \mathbf{X}))=\mathbf{U} \mathbf{\hat{G}} \mathbf{U}^T \mathbf{X},
\end{equation}
where $\odot$ denotes element-wise multiplication and $\mathbf{\hat{G}}=\text{diag}(\hat{g_1}, ..., \hat{g_n})$ denotes a diagonal matrix with $n$ spectral filter coefficients.

To reduce the current learning complexity of $\mathcal{O}(n)$ to that of conventional CNN, i.e., $\mathcal{O}(K)$, (\ref{eqn: graph convolution}) can be approximated using the $K$th order polynomials as follows
\begin{equation}
\label{eqn: poly expansion}
    \mathbf{U} \mathbf{\hat{G}} \mathbf{U}^T \mathbf{X} \approx \mathbf{U} (\sum_{i=0}^{K}\theta_i \mathbf{\Lambda}^i) \mathbf{U}^T \mathbf{X}= \sum_{i=0}^{K}\theta_i \mathbf{\hat{L}}^i \mathbf{X},
\end{equation}
where $\theta_i$ denotes learnable parameters. To further reduce computational cost, Defferrard \textit{et al.} \cite{defferrard2016convolutional} proposed to use Chebyshev polynomials to approximate the filtering operation as follows
\begin{equation}
\label{eqn: chebyshev}
    \mathbf{U} \mathbf{\hat{G}} \mathbf{U}^T \mathbf{X} = \sum_{i=0}^{K}\theta_i T_i(\mathbf{\hat{L}}^{'}) \mathbf{X},
\end{equation}
where $\mathbf{\hat{L}}^{'} = \frac{2}{\lambda_{max}}\mathbf{\hat{L}} - \mathbf{I}$ denotes the scaled normalized Laplacian with its eigenvalues lying within $[-1,1]$, $\lambda_{max}$ denotes the maximum eigenvalue of $\mathbf{\hat{L}}$, and $T_i(x)$ denotes the Chebyshev polynomials recursively defined as $T_i(x)=2xT_{i-1}(x) - T_{i-2}(x)$ with $T_0(x)=1$ and $T_1(x)=x$.

The GCN proposed in \cite{kipf2017semi} made a few approximations to simplify the filtering operation in (\ref{eqn: chebyshev}): 1)~use $K=1$; 2)~set $\lambda_{max}=2$; and 3)~set $\theta_1=-\theta_0$. The resulted GCN arrives at (\ref{eqn: gcn}). Essentially, the graph convolutional operations defined in (\ref{eqn: gcn}) and (\ref{eqn: sgc}) behave like a low-pass filter by smoothing the features of each node on the graph using node features in its neighborhood. 
 
\section{Regularized Graph Neural Network}
\label{sec: RGNN}
In this section, we present our regularized graph neural network (RGNN), specifically, the biologically inspired adjacency matrix, the dynamics of RGNN, and two regularizers, i.e., node-wise domain adversarial training (NodeDAT) and emotion-aware distribution learning (EmotionDL). 

\begin{figure*}[!t]
\centering
\includegraphics[width=\linewidth]{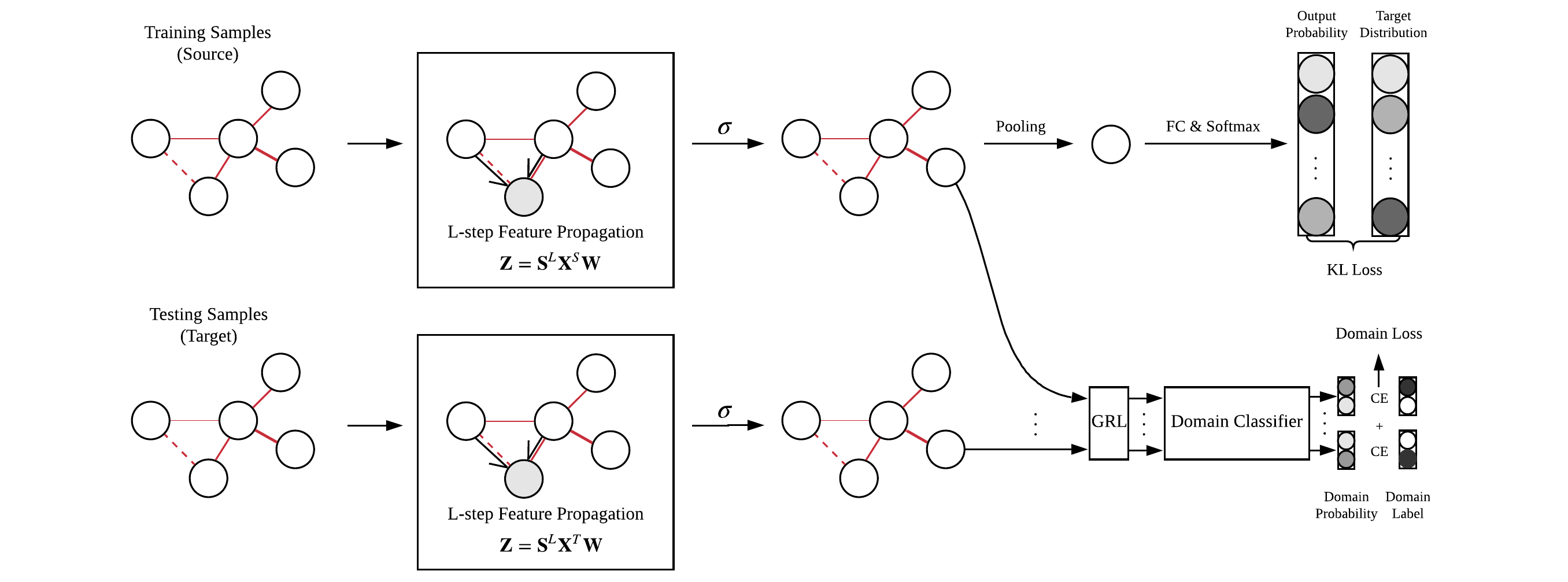}
\caption{The overall architecture of our RGNN model. FC denotes fully-connected layer. CE denotes cross-entropy loss. KL denotes Kullback-Leibler divergence \cite{kullback1951information}. GRL denotes gradient reversal layer \cite{ganin2016domain}.}
\label{fig: architecture}
\end{figure*}
\begin{figure}[!t]
\centering
\includegraphics[width=\linewidth]{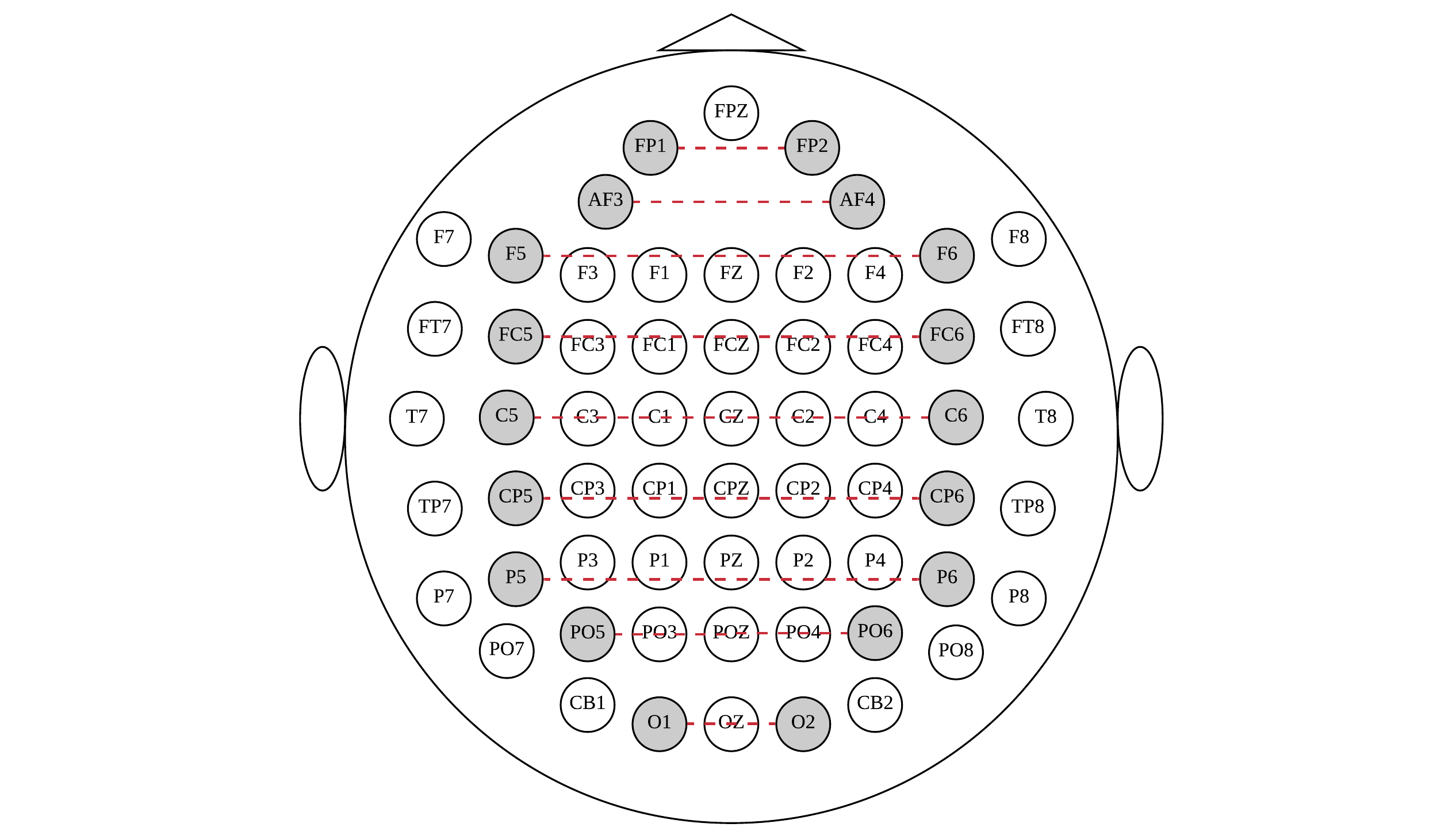}
\caption{The 62-channel EEG placement used to collect data in SEED and SEED-IV. Gray symmetric channels are connected globally via red dashed lines.}
\label{fig: 62 channel eeg}
\end{figure}

\subsection{Adjacency Matrix in RGNN}
\label{sec: adjacency matrix}
The adjacency matrix $\mathbf{A} \in \mathbb{R}^{n\times n}$ in RGNN represents the topological structure of EEG channels and is essential to graph representation learning, where $n$ denotes the number of channels in EEG signals. Each entry $\mathbf{A}_{ij}$ is learnable and indicates the weight of connection between channels $i$ and $j$. Note that $\mathbf{A}$ contains self-loops. To reduce overfitting, we model $\mathbf{A}$ as a symmetric matrix by using only $\frac{n(n+1)}{2}$ number of parameters instead of $n^2$. Salvador \textit{et al.} \cite{salvador2005neurophysiological} observed that the strength of connection between brain regions decays as an inverse square function of physical distance. Hence, we initialize the local inter-channel relations in our adjacency matrix as follows
\begin{equation}
\label{eqn: anatomical connection}
    \mathbf{A}_{ij} = \text{min}(1, \frac{\delta}{d_{ij}^2}),
\end{equation}
where $d_{ij}$, $i, j=1,2,...,n$, denotes the physical distance between channels $i$ and $j$, which is computed from their 3D coordinates obtained from the data sheet of the recording device, and $\delta > 0$ denotes a calibration constant. Achard and Bullmore \cite{achard2007efficiency} observed that sparse fMRI networks, comprising around 20\% of all possible connections, typically maximize the efficiency of the network topology. Therefore, we choose $\delta =5$ such that around 20\% of the entries in $\mathbf{A}$ are non-negligible. We empirically regard entries having values larger than $0.1$ as non-negligible connections. 

Bullmore and Sporns \cite{bullmore2012economy} suggested that the brain organization is shaped by an economic trade-off between minimizing wiring costs and network running costs. Minimizing wiring costs encourages local inter-channel connections as modelled in (\ref{eqn: anatomical connection}). However, minimizing network running costs encourages certain global inter-channel connections for high efficiency of information transfer across the network as a whole. To this end, we add several global connections to our adjacency matrix to improve the network efficiency. The global connections depend on specific electrode placement adopted in experiments. Fig. \ref{fig: 62 channel eeg} depicts the global connections in both SEED \cite{zheng2015investigating} and SEED-IV \cite{zheng2018emotionmeter}. The selection of global channels is supported by prior studies showing that the asymmetry in neuronal activities between the left and right hemispheres is informative in valence and arousal predictions \cite{schmidt2001frontal}. To leverage the differential asymmetry information, we initialize the global inter-channel relations in $\mathbf{A}$ to $[-1, 0]$ as follows
\begin{equation}
\label{eqn: global connection}
    \mathbf{A}_{ij} = \mathbf{A}_{ij} - 1,
\end{equation}
where $(i,j)$ denotes the indices of global channel pairs: (FP1, FP2), (AF3, AF4), (F5, F6), (FC5, FC6), (C5, C6), (CP5, CP6), (P5, P6), (PO5, PO6) and (O1, O2). Note that we select these indices because 1) they are connected to a large number of nodes in their immediate neighborhood, which maximizes the effects of EEG asymmetry; and 2) they empirically perform slightly better than alternative sets of indices (see Section \ref{sec: ablation study}). Our adjacency matrix $\mathbf{A}$ obtained in (\ref{eqn: anatomical connection}) and (\ref{eqn: global connection}) aims to represent the brain network which combines both local anatomical connectivity and emotion-related global functional connectivity. 

\subsection{Dynamics of RGNN}
\label{sec: rgnn}
Our RGNN model extends the SGC model \cite{wu2019simplifying}. The architecture of RGNN is illustrated in Fig. \ref{fig: architecture}. Given EEG features $\mathbf{X} \in \mathbb{R}^{N \times n \times d}$ and labels $\mathbf{Y} \in \mathbb{Z}^{N}$, where $N$ denotes the number of training samples, $\mathbf{Y}_i \in \{0,1,...,C-1\}$ denotes the class index, and $C$ denotes the number of classes. Our model aims to minimize the following cross-entropy loss:
\begin{equation}
\label{eqn: original loss function}
    \Phi = -\sum_{i=1}^{N} \text{log}(p(\mathbf{Y}_i|\mathbf{X}_i, \theta)) + \alpha ||\mathbf{A}||_1,
\end{equation}
where $\theta$ denotes the model parameters we want to optimize, and $\alpha$ denotes the strength of L1 sparse regularization for our adjacency matrix $\mathbf{A}$. 

By passing each feature matrix $\mathbf{X}_i$ into our RGNN, the output probability of class $\mathbf{Y}_i$ can be computed as follows
\begin{equation}
\label{eqn: output probability}
\begin{split}
    \mathbf{Z_i} &=\mathbf{S}^L \mathbf{X_i} \mathbf{W},\\
    p(\mathbf{Y}_i|\mathbf{X}_i, \theta) &= \text{softmax}_{\mathbf{Y}_i}(\text{pool}(\sigma(\mathbf{Z_i}))\mathbf{W}^O),
\end{split}
\end{equation}
where $\mathbf{S} \in \mathbb{R}^{n \times n}$, $\mathbf{W} \in \mathbb{R}^{d \times d^{'}}$ and $L$ follow the definitions in (\ref{eqn: sgc}), $\sigma (x) = \text{max}(0,x)$ denotes a non-linear transformation, $\mathbf{W}^O \in \mathbb{R}^{d^{'} \times C}$ denotes the output weight matrix, and $\text{pool}(\cdot)$ denotes the sum pooling across all nodes on the graph. We choose sum pooling because it demonstrated more expressive power than mean pooling and max pooling \cite{xu2018how}. Note that we use the absolute values of $\mathbf{A}$ to compute the degree matrix $\mathbf{D}$ (see (\ref{eqn: gcn})) because $\mathbf{A}$ has negative entries, e.g., global connections. 

\subsubsection{Node-wise Domain Adversarial Training}
\label{sec: NodeDAT}
EEG signals vary significantly across different subjects, which hinders the generalizability of trained classifiers in subject-independent classification settings. To improve the robustness of our model across subjects, we extend the domain adversarial training \cite{ganin2016domain} by proposing a node-wise domain adversarial training (NodeDAT) method to reduce the discrepancies between source and target domains, i.e., training and testing sets, respectively. Specifically, a domain classifier is proposed to classify each node representation into either source domain or target domain. During optimization, our model aims to confuse the domain classifier by learning domain-invariant representations. Compared to \cite{ganin2016domain}, which only regularizes the pooled representation in the last layer, our NodeDAT method has finer-grained regularization because it explicitly regularizes each node representation before pooling. 

Specifically, given labelled source/training data $\mathbf{X}^S \in \mathbb{R}^{N \times n \times d}$ (in this subsection, we denote $\mathbf{X}$ by $\mathbf{X}^S$ for better clarity) and unlabelled target/testing data $\mathbf{X}^T \in \mathbb{R}^{N \times n \times d}$, where in practice $\mathbf{X}^T$ can be either oversampled or donwsampled to have the same number of samples as $\mathbf{X}^S$ \cite{ganin2016domain}, the domain classifier aims to minimize the sum of the following two binary cross-entropy losses:
\begin{equation}
\label{eqn: domain loss function}
    \Phi_D = -\sum_{i=1}^{N}\sum_{j=1}^{n} (\text{log}(p_j(0|\mathbf{X}^S_i, \theta_D)) + \text{log}(p_j(1|\mathbf{X}^T_i, \theta_D))),
\end{equation}
where $\theta_D$ denotes the parameters of the domain classifier, $0$ and $1$ denote source and target domains, respectively. Intuitively, the domain classifier is learned to classify source data as 0 and target data as 1. The domain probabilities $p_j(\cdot)$ for the $j$th node on the $i$th example are computed as
\begin{equation}
\label{eqn: domain probability}
\begin{split}
    p_j(0|\mathbf{X}^{S}_i, \theta_D) = \text{softmax}_0(\sigma(\mathbf{Z}^S_{ij}) \mathbf{W}^D),\\
    p_j(1|\mathbf{X}^{T}_i, \theta_D) = \text{softmax}_1(\sigma(\mathbf{Z}^T_{ij}) \mathbf{W}^D),
\end{split}
\end{equation}
where $\mathbf{Z}^{\{S,T\}}_{ij}$ denotes the $j$th node representation in $\mathbf{Z}^{\{S,T\}}_i$, and $\mathbf{W}^D \in \mathbb{R}^{d^{'} \times 2}$ denotes the matrix parameter in the domain classifier, i.e., $\theta_D$. 

In order to confuse the domain classifier and learn domain invariant node presentation $\mathbf{Z}^{\{S,T\}}_{ij}$, we implement a gradient reversal layer (GRL) \cite{ganin2016domain} that acts like an identity layer in the forward propagation and reverses the gradients of the domain classifier during backpropagation. Consequently, the parameters in the feature extractor essentially perform gradient ascent with respect to the gradients from the domain classifier. The reversed gradients are further scaled by a GRL scaling factor $\beta$ which gradually increases from 0 to 1 as the training progresses. The gradually increasing $\beta$ allows our domain classifier to be less sensitive to noisy inputs at the early stages of the training process. Specifically, as suggested in \cite{ganin2016domain}, we let $\beta = \frac{2}{1 + e^{-10p}} - 1$, where $p \in [0,1]$ denotes the progression of training.

\subsubsection{Emotion-aware Distribution Learning}
\label{sec: emotiondl}
Participants may not always generate the intended emotions when watching emotion-eliciting stimuli, which may have negative impact on model performance \cite{zhu2004class}. To this end, we propose an emotion-aware distribution learning (EmotionDL) method to learn a distribution of classes instead of one single class for each training sample. Specifically, we convert each training label $\mathbf{Y}_i \in \{0,1,...,C-1\}$ into a prior probability distribution of all classes $\hat{\mathbf{Y}}_i \in \mathbb{R}^{C}$, where $\hat{\mathbf{Y}}_{ic}$ denotes the probability of class c in $\hat{\mathbf{Y}}_i$. The conversion is dataset-dependent. SEED has three classes: negative, neutral, and positive with corresponding class indices 0, 1, and 2, respectively. We convert $\mathbf{Y}$ as follows

\begin{equation}
\label{eqn: convert SEED label}
\hat{\mathbf{Y}}_i = 
\begin{cases} 
  (1 - \frac{2\epsilon}{3}, \frac{2\epsilon}{3}, 0),& \mathbf{Y}_i = 0, \\
  (\frac{\epsilon}{3}, 1 - \frac{2\epsilon}{3}, \frac{\epsilon}{3}),& \mathbf{Y}_i = 1, \\
  (0, \frac{2\epsilon}{3}, 1 - \frac{2\epsilon}{3}),& \mathbf{Y}_i = 2 ,
\end{cases}
\end{equation}
where $\epsilon \in [0,1]$ denotes a hyper-parameter controlling the noise level in the training labels. This conversion mechanism is based on our assumption that participants are unlikely to generate opposite emotions when watching emotion-eliciting stimuli. Therefore, for each class, the converted class distribution centers on the original class and has zero probabilities at its opposite classes. 

SEED-IV has four classes: neutral, sad, fear, and happy with corresponding class indices 0, 1, 2, and 3, respectively. We convert $\mathbf{Y}$ as follows
\begin{equation}
\label{eqn: convert SEED-IV label}
\hat{\mathbf{Y}}_i = 
\begin{cases} 
  (1 - \frac{3\epsilon}{4}, \frac{\epsilon}{4}, \frac{\epsilon}{4}, \frac{\epsilon}{4}), & \mathbf{Y}_i = 0, \\
  (\frac{\epsilon}{3}, 1 - \frac{2\epsilon}{3}, \frac{\epsilon}{3}, 0), & \mathbf{Y}_i = 1, \\
  (\frac{\epsilon}{4}, \frac{\epsilon}{4}, 1 - \frac{3\epsilon}{4}, \frac{\epsilon}{4}), & \mathbf{Y}_i = 2, \\
  (\frac{\epsilon}{3}, 0, \frac{\epsilon}{3}, 1 - \frac{2\epsilon}{3}), & \mathbf{Y}_i = 3.
\end{cases}
\end{equation}
This conversion is based on the distances between the four emotion classes on the valence-arousal plane. Specifically, in the self-reported ratings \cite{zheng2018emotionmeter} for SEED-IV, neutral, sad, fear, and happy movie ratings cluster in the zero valence zero arousal, low valence low arousal, low valence high arousal, and high valence high arousal regions, respectively. We assume that participants are likely to generate emotions that have similar ratings in either valence or arousal dimensions, e.g., both angry and happy have high arousal, but unlikely to generate emotions that are far away in both dimensions, e.g., sad and happy are different in both valence and arousal dimensions.

After obtaining the converted class distributions $\hat{\mathbf{Y}}$, our model can be optimized by minimizing the following Kullback-Leibler (KL) divergence \cite{kullback1951information} instead of (\ref{eqn: original loss function}):
\begin{equation}
\label{eqn: kl loss function}
    \Phi^{'} = \sum_{i=1}^{N} \text{KL}(p(\mathbf{Y}|\mathbf{X}_i, \theta), \hat{\mathbf{Y}}_i) + \alpha ||\mathbf{A}||_1,
\end{equation}
where $p(\mathbf{Y}|\mathbf{X}_i, \theta)$ denotes the output probability distribution computed via (\ref{eqn: output probability}). Note that EmotionDL incorporates more prior knowledge than label smoothing, which simply adds uniform noise to other classes.

\subsubsection{Optimization of RGNN}
\label{sec: optimization}
Combining both NodeDAT and EmotionDL, the overall loss function $\Phi^{''}$ of RGNN is computed as follows
\begin{equation}
\label{eqn: overall loss function}
    \Phi^{''} = \Phi^{'} + \Phi_D.
\end{equation}
The detailed algorithm for training RGNN is presented in Algorithm \ref{algorithm}.

\begin{algorithm}
\caption{The Training Algorithm of RGNN}
\label{algorithm}
\begin{algorithmic}[1]
\Require Training samples $\mathbf{X}$ and $\hat{\mathbf{Y}}$, unlabelled testing samples $\mathbf{X}^T$, learning rate $\eta$, number of epochs $T$, batch size $B$, other regularization hyper-parameters; 
\Ensure The learned model parameters in RGNN;
\State{Randomly initialize model parameters in RGNN using Xavier initialization \cite{glorot2010understanding};}
\State{Initialize adjacency matrix $\mathbf{A}$ based on (\ref{eqn: anatomical connection}) and (\ref{eqn: global connection});}
\For{$i=1$: $T$}
\Repeat
\State{Draw one batch of training samples $\mathbf{X}_B$ and $\hat{\mathbf{Y}}_B$ \hspace*{10mm}from $\mathbf{X}$ and $\hat{\mathbf{Y}}$, respectively;}
\State{Draw one batch of testing samples $\mathbf{X}_B^T$ from $\mathbf{X}^T$;}
\State{Compute degree matrix $\mathbf{D}$ based on (\ref{eqn: gcn});}
\State{Compute normalized adjacency matrix $\mathbf{S}$ based \hspace*{10mm}on (\ref{eqn: sgc});}
\State{Compute output representation $\mathbf{Z}$ based on (\ref{eqn: output probability});}
\State{Use $\mathbf{X}_B$ and $\hat{\mathbf{Y}}_B$ to compute KL loss $\Phi^{'}$ based on  \hspace*{10mm}(\ref{eqn: kl loss function});}
\State{Use $\mathbf{X}_B$ and $\mathbf{X}_B^T$ to compute domain loss $\Phi_D$ \hspace*{10mm}based on (\ref{eqn: domain loss function});}
\State{Compute GRL scaling factor $\beta$;}
\State{Update $\mathbf{W}^D \leftarrow \mathbf{W}^D - \eta\frac{\partial \Phi_D}{\partial \mathbf{W}^D}$;}
\State{Update $\mathbf{W}^O \leftarrow \mathbf{W}^O - \eta\frac{\partial \Phi^{'}}{\partial \mathbf{W}^O}$;}
\State{Update $\mathbf{W} \leftarrow \mathbf{W} - \eta(\frac{\partial \Phi^{'}}{\partial \mathbf{W}} - \beta\frac{\partial \Phi_D}{\partial \mathbf{W}})$;}
\State{Update $\mathbf{A} \leftarrow \mathbf{A} - \eta(\frac{\partial \Phi^{'}}{\partial \mathbf{A}} - \beta\frac{\partial \Phi_D}{\partial \mathbf{A}})$;}
\Until{all samples in $\mathbf{X}$ have been drawn;}
\EndFor
\end{algorithmic}
\end{algorithm}

\section{Experimental Settings}
\label{sec: experimental settings}
In this section, we present the datasets, classification settings and model settings in our experiments.
\subsection{Datasets}
We conduct experiments on two public datasets, namely SEED and SEED-IV. The SEED dataset \cite{zheng2015investigating} comprises EEG data of 15 subjects (7 males) recorded in 62 channels using the ESI NeuroScan System\footnote{https://compumedicsneuroscan.com/}. The data were collected when participants watch emotion-eliciting movies in three types of emotions, namely negative, neutral and positive. Each movie lasts around 4 minutes. Three sessions of data are collected and each session comprises 15 trials/movies for each subject. To make a fair comparison with existing studies, we directly use the pre-computed differential entropy (DE) features smoothed by linear dynamic systems (LDS) \cite{zheng2015investigating} in SEED. DE extends the idea of Shannon entropy and measures the complexity of a continuous random variable. 
In SEED, DE features are pre-computed over five frequency bands (delta, theta, alpha, beta and gamma) for each second of EEG signals (without overlapping) in each channel.

The SEED-IV dataset \cite{zheng2018emotionmeter} comprises EEG data of 15 subjects (7 males) recorded in 62 channels\footnote{SEED-IV also contains eye movement data, which we do not use in our experiments.}. The recording device is the same as the one used in SEED. The data were collected when participants watch emotion-eliciting movies in four types of emotions, namely neutral, sad, fear, and happy. Each movie lasts around 2 minutes. Three sessions of data are collected and each session comprises 24 trials/movies for each subject. Similar to SEED, we adopt the pre-computed DE features from SEED-IV.

\subsection{Classification Settings}
We closely follow prior studies to conduct both subject-dependent and subject-independent classifications on both SEED and SEED-IV to evaluate our model.
\subsubsection{Subject-Dependent Classification}
For SEED, we follow the experimental settings in \cite{zheng2015investigating, song2018eeg, li2018bi} to evaluate our RGNN model using subject-dependent classification. Specifically, for each subject, we train our model using the first 9 trials as the training set and the remaining 6 trials as the testing set. We evaluate the model performance by using the accuracy averaged across all subjects over two sessions of EEG data\cite{zheng2015investigating}. 
Similarly, for subject-dependent classification on SEED-IV, we follow the experimental settings in \cite{zheng2018emotionmeter, li2019novel} to use the first 16 trials for training and the remaining 8 trials containing all emotions (two trials per emotion class) for testing. We evaluate our model using data from all three sessions \cite{zheng2018emotionmeter}. 

\subsubsection{Subject-Independent Classification}
For SEED, we follow the experimental settings in \cite{zheng2016personalizing, song2018eeg, li2018bi} to evaluate our RGNN model using subject-independent classification. Specifically, we adopt leave-one-subject-out cross-validation, i.e, during each fold, we train our model on 14 subjects and test on the remaining subject. We evaluate the model performance using the accuracy averaged cross all test subjects over one session of EEG data \cite{zheng2016personalizing}. 
Similarly, for SEED-IV, we follow the experimental settings in \cite{li2019novel} to evaluate our RGNN model using subject-independent classification. We evaluate our model using data from all three sessions \cite{li2019novel}. 
\begin{table*}[!t]
\centering
\caption{Subject-dependent classification accuracy (mean/std) on SEED and SEED-IV}
\begin{tabular}{c|c|c|c|c|c|c|c}
\hline
 & \multicolumn{6}{c|}{\textbf{SEED}} & \textbf{SEED-IV} \\
\hline\hline
\textbf{Model} & delta band & theta band & alpha band & beta band & gamma band & \textbf{all bands} & \textbf{all bands}\\
\hline\hline
SVM & 60.50/14.14 & 60.95/10.20 & 66.64/14.41 & 80.76/11.56 & 79.56/11.38 & 83.99/09.92 & 56.61/20.05\\
\hline
GSCCA \cite{zheng2016multichannel} & 63.92/11.16 & 64.64/10.33 & 70.10/14.76 & 76.93/11.00 & 77.98/10.72 & 82.96/09.95 & 69.08/16.66\\ 
\hline
DBN \cite{zheng2015investigating} & 64.32/12.45 & 60.77/10.42 & 64.01/15.97 & 78.92/12.48 & 79.19/14.58 & 86.08/08.34 & 66.77/07.38\\ 
\hline
STRNN \cite{zhang2018spatial} & \textbf{80.90}/12.27 & \textbf{83.35}/09.15 & \textbf{82.69}/12.99 & 83.41/10.16 & 69.61/15.65 & 89.50/07.63 & -\\ 
\hline
DGCNN \cite{song2018eeg} & 74.25/11.42 & 71.52/05.99 & 74.43/12.16 & 83.65/10.17 & 85.73/10.64 & 90.40/08.49 & 69.88/16.29\\ 
\hline
BiDANN \cite{li2018bi} & 76.97/10.95 & 75.56/07.88 & 81.03/11.74 & \textbf{89.65}/09.59 & 88.64/09.46 & 92.38/07.04 & 70.29/12.63\\ 
\hline
EmotionMeter \cite{zheng2018emotionmeter} & - & - & - & - & - & - & 70.58/17.01\\ 
\hline
BiHDM \cite{li2019novel} (SOTA)& - & - & - & - & - & 93.12/06.06 & 74.35/14.09\\ 
\hline
RGNN (Our model) & 76.17/07.91 & 72.26/07.25 & 75.33/08.85 & 84.25/12.54 & \textbf{89.23}/08.90 & \textbf{94.24}/05.95& \textbf{79.37}/10.54\\ 
\hline
\end{tabular}
\label{table: subject-dependent}
\end{table*}

\begin{table*}[!t]
\centering
\caption{Subject-independent classification accuracy (mean/std) on SEED and SEED-IV}
\begin{tabular}{c|c|c|c|c|c|c|c}
\hline
 & \multicolumn{6}{c|}{\textbf{SEED}} & \textbf{SEED-IV} \\
\hline\hline
\textbf{Model} & delta band & theta band & alpha band & beta band & gamma band & \textbf{all bands} & \textbf{all bands} \\
\hline\hline
SVM & 43.06/08.27 & 40.07/06.50 & 43.97/10.89 & 48.63/10.29 & 51.59/11.83 & 56.73/16.29 & 37.99/12.52\\
\hline
TCA \cite{pan2010domain} & 44.10/08.22 & 41.26/09.21 & 42.93/14.33 & 43.93/10.06 & 48.43/09.73 & 63.64/14.88 & 56.56/13.77\\ 
\hline
SA \cite{fernando2013unsupervised} & 53.23/07.47 & 50.60/08.31 & 55.06/10.60 & 56.72/10.78 & 64.47/14.96 & 69.00/10.89 & 64.44/09.46\\ 
\hline
T-SVM \cite{collobert2006large} & - & - & - & - & - & 72.53/14.00 & -\\ 
\hline
DGCNN \cite{song2018eeg} & 49.79/10.94 & 46.36/12.06 & 48.29/12.28 & 56.15/14.01 & 54.87/17.53 & 79.95/09.02 & 52.82/09.23\\ 
\hline
DAN \cite{li2018cross} & - & - & - & - & - & 83.81/08.56 & 58.87/08.13\\ 
\hline
BiDANN-S \cite{li2018bi} & 63.01/07.49 & \textbf{63.22}/07.52 & \textbf{63.50}/09.50 & 73.59/09.12 & 73.72/08.67 & 84.14/06.87 & 65.59/10.39\\ 
\hline
BiHDM \cite{li2019novel} (SOTA) & - & - & - & - & - & \textbf{85.40}/07.53 & 69.03/08.66\\ 
\hline
RGNN (Our model) & \textbf{64.88}/06.87 & 60.69/05.79 & 60.84/07.57 & \textbf{74.96}/08.94 & \textbf{77.50}/08.10 & 85.30/06.72 & \textbf{73.84}/08.02\\ 
\hline
\end{tabular}
\label{table: subject-independent}
\end{table*}
\subsection{Model Settings in RGNN}
For hyper-parameters of RGNN in all experiments, we empirically set the number of convolutional layers $L=2$, dropout rate of $0.7$ at the output fully-connected layer \cite{srivastava2014dropout}, and batch size of $16$. We use Adam \cite{kingma2014adam} to optimize model parameters using gradient descent. We only tune the output feature dimension $d^{'}$, label noise level $\epsilon$, learning rate $\eta$, L1 regularization factor $\alpha$, and L2 regularization for each experiment. Note that we only adopt NodeDAT in subject-independent classification experiments. Our model is publicly available\footnote{https://github.com/zhongpeixiang/RGNN}. We compare our model with several baselines, which are all cited from published results \cite{song2018eeg, li2018bi, zhang2018spatial, li2019novel}.

\section{Performance Evaluations}
\label{sec: performance comparisons with baselines}
In this section, we present model evaluation results and investigate the critical frequency bands and confusion matrices of our RGNN model.

\subsection{Subject-Dependent Classification}
Table \ref{table: subject-dependent} presents the subject-dependent classification accuracy of our RGNN model and all baselines on both SEED and SEED-IV. The performance on SEED in the individual delta, theta, alpha, beta, and gamma bands is reported as well. It is encouraging to see that our model achieves better performance than all baselines including the state-of-the-art BiHDM on both datasets when features from all frequency bands are used. 
In particular, our model performs better than DGCNN, another GNN-based model that leverages the topological structure of EEG channels. Besides the proposed two regularizers (see Table~\ref{table: ablation study}), the main performance improvement can be attributed to two factors: 1)~our adjacency matrix incorporates the emotion-discriminative global inter-channel asymmetry relation between the left and right hemispheres; and 2)~our model has less concern of overfitting by extending SGC, which is much simpler than ChebNet \cite{defferrard2016convolutional} used in DGCNN. 

\subsection{Subject-Independent Classification}
Similar to Table \ref{table: subject-dependent}, Table \ref{table: subject-independent} presents the subject-independent classification results. When using features from all frequency bands, our model performs marginally worse than BiHDM on SEED but much better than BiHDM on SEED-IV (nearly 5\% improvement). In addition, our model achieves the lowest standard deviation in accuracy compared to all baselines on both datasets, showing the robustness of our model against cross-subject variations. 

Comparing the results shown in Tables \ref{table: subject-dependent} and \ref{table: subject-independent}, we find that the accuracy obtained in subject-independent settings is consistently worse than the accuracy obtained in subject-dependent settings by around 5\% to 30\% for every model. This finding is unsurprising because the variability of EEG signals across subjects makes subject-independent classification more challenging. However, an interesting observation is that the performance gap between these two settings is gradually decreasing from around 27\% on SEED and 19\% on SEED-IV using SVM to around 9\% on SEED and 6\% on SEED-IV using our model. One possible reason for the diminishing performance gap is that recent deep learning models in subject-independent classification settings are becoming better at leveraging a large amount of data and learning subject-invariant EEG representations. This observation seems to indicate that transfer learning may be a necessary tool for emotion recognition in cross-subject settings. 

\subsection{Performance Comparison of Frequency Bands}
We further compare the performance of our model and all baselines on SEED using features from different frequency bands, as reported in Tables \ref{table: subject-dependent} and \ref{table: subject-independent}. In subject-dependent experiments, STRNN achieves the highest accuracy in delta, theta and alpha bands, BiDANN performs best in beta band, and our model performs best in gamma band. In subject-independent experiments, BiDANN-S achieves the highest accuracy in theta and alpha bands, and our model performs best in delta, beta and gamma bands. 

We investigate the critical frequency bands for emotion recognition. For both subject-dependent and subject-independent settings on SEED, we compare the performance of each model across different frequency bands. In general, most models including ours achieve better performance on beta and gamma bands than delta, theta and alpha bands, with one exception of STRNN, which performs the worst on gamma band. This observation is consistent with the literature \cite{ray1985eeg, zheng2015investigating}. 
One subtle difference between our model and other models is that our model performs consistently better in gamma band than beta band, whereas other models perform comparably in both bands, indicating that gamma band may be the most discriminative band for our model.

\subsection{Confusion Matrix}
\begin{figure}[!t]
\centering
\includegraphics[width=\linewidth]{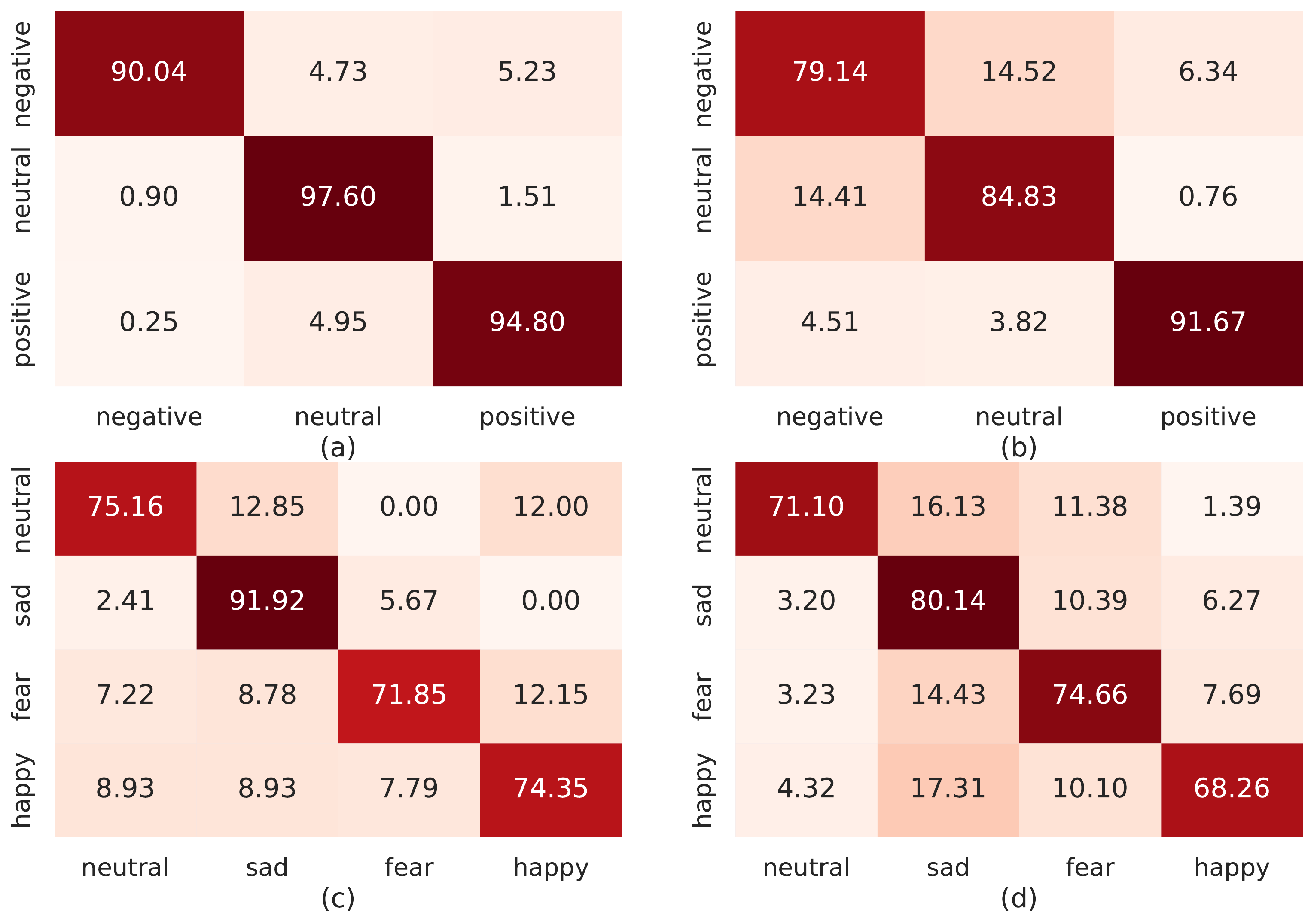}
\caption{Confusion matrices of RGNN. (a) Subject-dependent classification on SEED. (b) Subject-independent classification on SEED. (c) Subject-dependent classification on SEED-IV. (d) Subject-independent classification on SEED-IV.}
\label{fig: CM}
\end{figure}
We present the confusion matrices of our model in Fig.~\ref{fig: CM}. For SEED, our model can recognize positive and neutral emotions better than negative emotion in both classification settings. Comparing subject-independent classification (see Fig.~\ref{fig: CM}(b)) to subject-dependent classification (see Fig.~\ref{fig: CM}(a)), the performance of our model gets relatively much worse at detecting negative emotion, indicating that participants are likely to generate distinct EEG patterns when experiencing negative emotion.

For SEED-IV, our model performs significantly better on sad emotion than all other emotions in both classification settings. Comparing subject-independent classification (see Fig.~\ref{fig: CM}(d)) to subject-dependent classification (see Fig.~\ref{fig: CM}(c)), the performance of our model gets relatively much worse at detecting sad emotion, which is similar to SEED. We note that fear is the only emotion that performs better in subject-independent classification than in subject-dependent classification. This finding indicates that participants watching horror movies may generate similar EEG patterns. 

\section{Discussion}
In this section, we conduct ablation study and sensitivity analysis for our RGNN model. We also analyze important brain regions and inter-channel relations for emotion recognition.
\subsection{Ablation Study}
\label{sec: ablation study}

\begin{table}[!t]
\centering
\caption{Ablation study for subject-independent classification accuracy (mean/std) on SEED and SEED-IV. Symbol ``$-$" indicates the following component is removed.}
\begin{tabular}{c|c|c}
\hline
\textbf{Model} & \textbf{SEED} & \textbf{SEED-IV} \\
\hline\hline
RGNN & \textbf{85.30}/06.72 & \textbf{73.84}/08.02\\ 
\hline\hline
correlation-based adjacency matrix & 84.41/06.94 & 72.73/08.36\\ 
\hline
coherence-based adjacency matrix & 84.02/07.05 & 72.26/08.48\\ 
\hline
random adjacency matrix & 83.57/07.34 & 71.78/08.64\\ 
\hline
$-$ symmetric adjacency matrix& 83.69/07.92 & 72.02/08.66\\ 
\hline\hline
$-$ global connection & 82.42/08.24 & 71.13/08.78\\ 
\hline
global connection alternative 1 & 84.52/06.87 & 73.29/08.18\\ 
\hline
global connection alternative 2 & 84.23/07.04 & 73.08/08.35\\
\hline\hline
$-$ NodeDAT & 81.92/09.35 & 71.65/09.43\\ 
\hline
DAT & 83.51/08.11 & 72.40/08.54\\ 
\hline\hline
$-$ EmotionDL & 82.27/08.81 & 70.76/09.22\\ 
\hline
\end{tabular}
\label{table: ablation study}
\end{table}

\begin{figure}[!t]
\centering
\includegraphics[width=\linewidth]{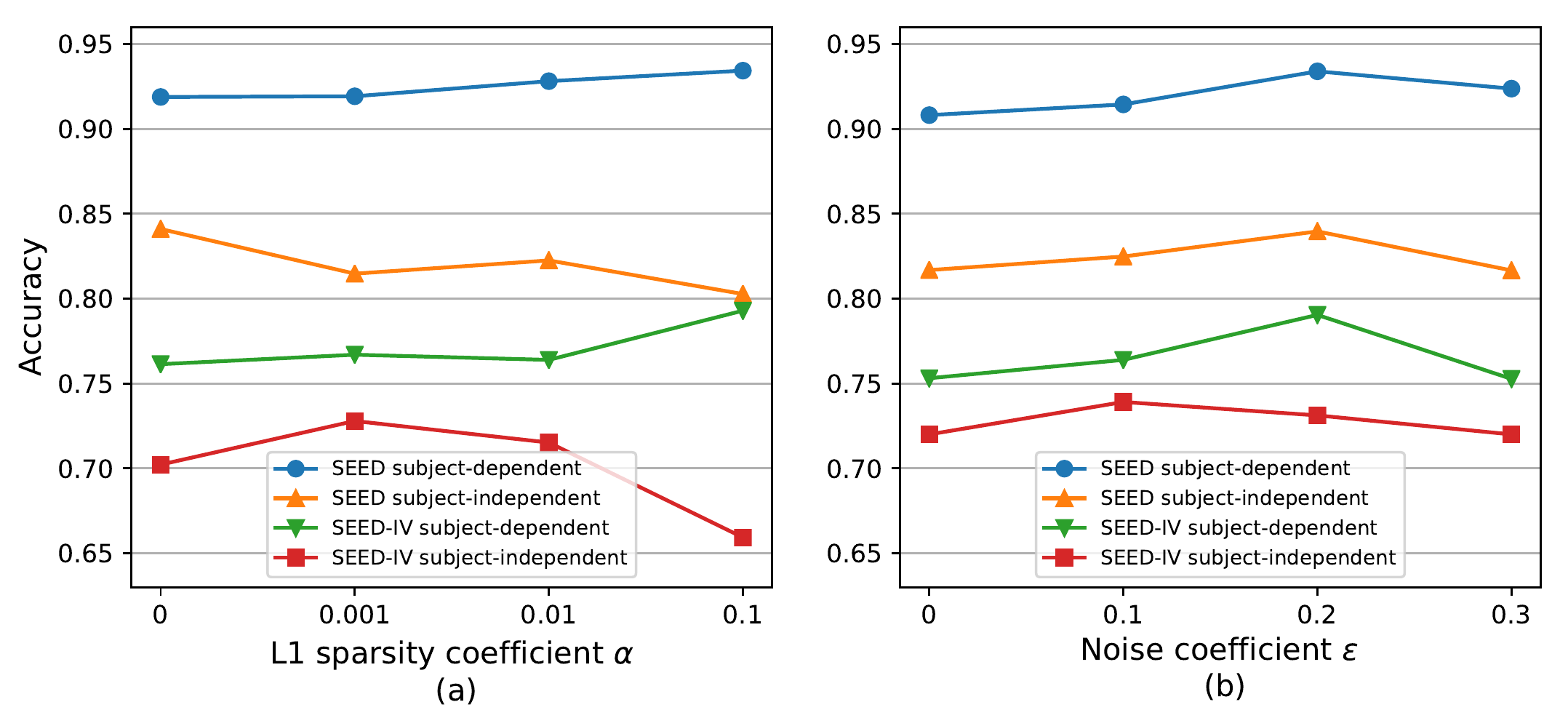}
\caption{Classification accuracy of RGNN with varying hyper-parameters. (a) L1 sparsity coefficient $\alpha$ in (\ref{eqn: original loss function}). (b) Noise coefficient $\epsilon$ in (\ref{eqn: convert SEED label}) and (\ref{eqn: convert SEED-IV label}).}
\label{fig: coefficients}
\end{figure}

\begin{figure*}[!t]
\centering
\includegraphics[width=\linewidth]{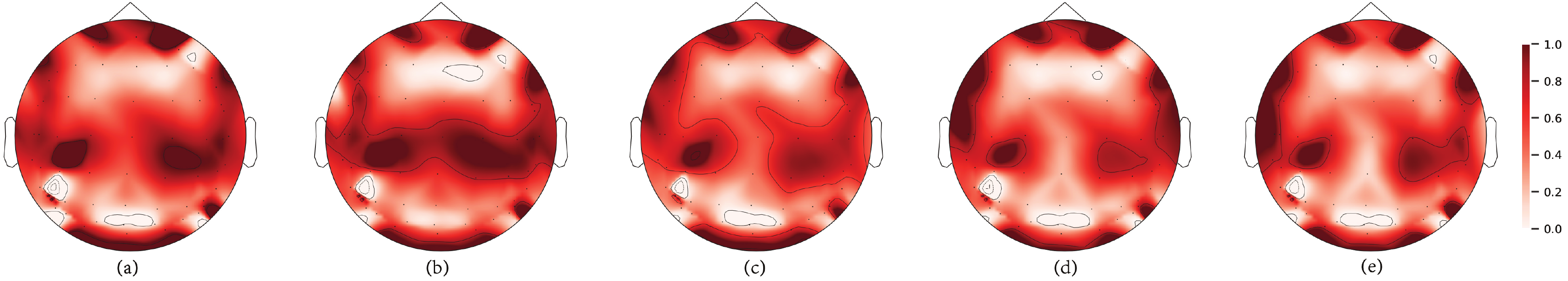}
\caption{Activation maps learned from subject-dependent classification on SEED-IV. (a) Delta band. (b) Theta band. (c) Alpha band. (d) Beta band. (e) Gamma band.}
\label{fig: activation maps}
\end{figure*}
We conduct ablation study to investigate the contribution of each key component in our model. Table \ref{table: ablation study} reports the subject-independent classification results on both datasets. We compared different initialization methods of the adjacency matrix and found that our distance-based method (see (\ref{eqn: anatomical connection})) obtains slightly better performance than functional connectivity-based methods, i.e., correlation and coherence computed from the training dataset. The uniformly randomly initialized adjacency matrix in $[0,1]$ performs worst, indicating that properly initializing the adjacency matrix is beneficial to model performance. Our symmetric adjacency matrix design also proves to be useful in reducing overfitting and improving accuracy. 

Removing the global connection causes noticeable performance drop on both datasets, demonstrating the importance of global connections in modelling the EEG differential asymmetry. Moreover, we compared the performance of alternative sets of global connections. Alternative 1 has global indices that are nearer to the central region, i.e., (FP1, FP2), (AF3, AF4), (F3, F4), (FC3, FC4), (C3, C4), (CP3, CP4), (P3, P4), (PO5, PO6) and (O1, O2). Alternative 2 has global indices that are further from the central region, i.e., (FP1, FP2), (AF3, AF4), (F7, F8), (FT7, FT8), (T7, T8), (TP7, TP8), (P7, P8), (PO7, PO8) and (O1, O2). Both alternatives perform slightly worse than our model but much better than no global connection, indicating that they are able to model EEG asymmetry to a certain extent.

Our NodeDAT regularizer has a noticeable positive impact on the performance of our model, suggesting that domain adaptation is helpful in cross-subject classification. To further investigate the impact of our node-level domain classifier, we experimented with replacing NodeDAT with a generic domain classifier DAT \cite{ganin2016domain}. The clear performance gap between DAT and our RGNN model indicates that NodeDAT can better regularize the model by learning subject-invariant representation at node level than graph level. In addition, if NodeDAT is removed, the performance of our model has a greater variance, validating the importance of our NodeDAT regularizer in improving the robustness of RGNN against cross-subject variations.

Our EmotionDL regularizer improves the performance of our model by around 3\% in accuracy on both datasets. This performance gain validates our assumption that participants are not always generating the intended emotions when watching emotion-eliciting stimuli. In addition, our EmotionDL regularizer can be easily adopted by other deep learning based emotion recognition models.

\subsection{Sensitivity Analysis}
We analyze the performance of our model across varying L1 sparsity coefficient $\alpha$ (see (\ref{eqn: original loss function})) and noise coefficient $\epsilon$ in EmotionDL (see (\ref{eqn: convert SEED label}) and (\ref{eqn: convert SEED-IV label})), as illustrated in Fig.~\ref{fig: coefficients}. For subject-dependent classification, increasing $\alpha$ from 0 to 0.1 generally increases the model performance. However, for subject-independent classification, increasing $\alpha$ beyond a certain threshold, i.e, 0.01 in Fig. \ref{fig: coefficients}(a), decreases the model performance. One possible explanation for the difference in model behaviors is that there is much less training data in subject-dependent classification, which thus requires a stronger regularization to reduce overfitting, whereas for subject-independent classification where the amount of training data is less of a concern, adding stronger regularization may introduce bias and hinder the learning efficacy. 

As illustrated in Fig. \ref{fig: coefficients}(b), our model behaves consistently across different experimental settings with varying noise coefficient $\epsilon$. Specifically, by increasing $\epsilon$, the performance of our model first increases and then decreases. In particular, our model usually performs best when $\epsilon$ is set to 0.2, demonstrating the existence of label noises and the necessity of addressing them on both datasets. Introducing excessive noise in EmotionDL causes performance drop, which is expected because excessive noise weakens the true learning signals.

\begin{figure}[!t]
\includegraphics[width=\linewidth]{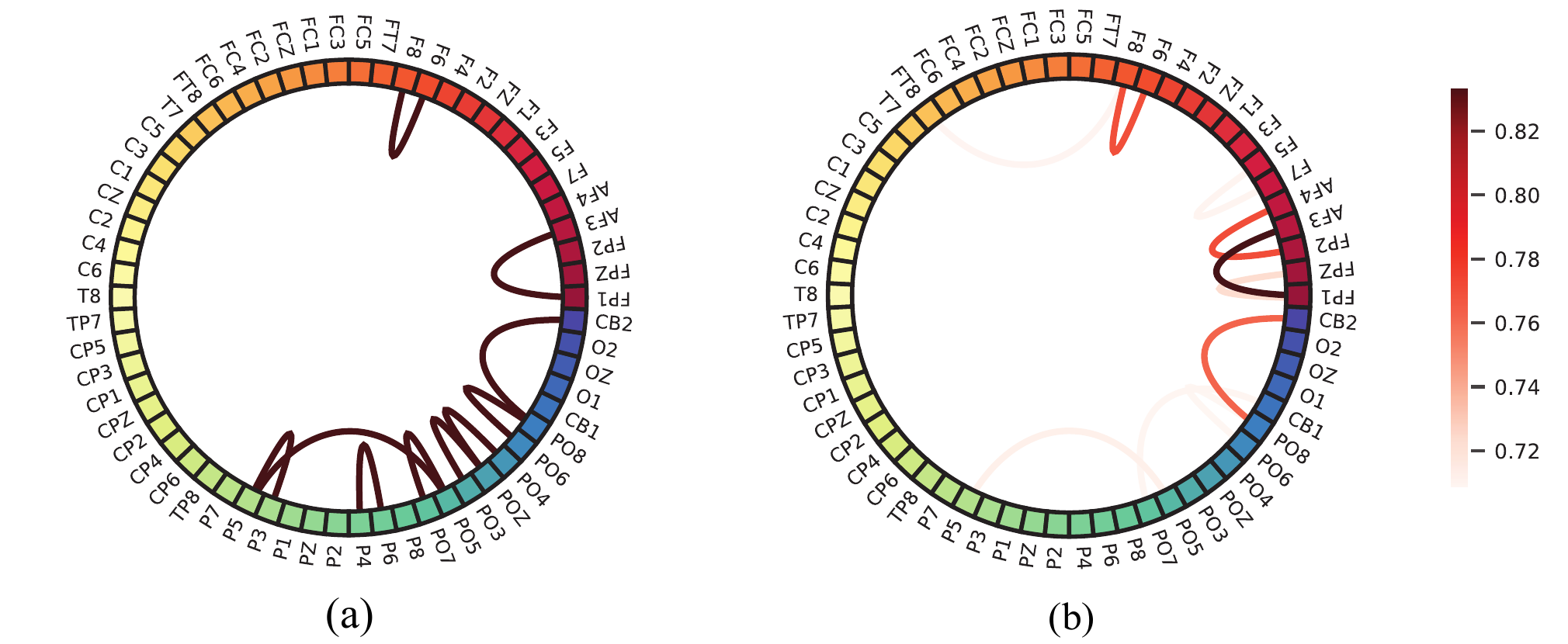}
\caption{Top 10 connections between channels in the adjacency matrix $\mathbf{A}$, excluding global connections in (\ref{eqn: global connection}) for better clarity. (a) Initialized $\mathbf{A}$ according to (\ref{eqn: anatomical connection}). (b) Learned and averaged $\mathbf{A}$ across five frequency bands in subject-dependent classification on both SEED and SEED-IV.}
\label{fig: electrode connectivity}
\end{figure}

\subsection{Analysis of Important Brain Regions and Inter-channel Relations}
We identify important brain regions for emotion recognition. Fig. \ref{fig: activation maps} shows the heatmaps of the diagonal elements in our learned adjacency matrix $\mathbf{A}$ in subject-dependent classification on SEED-IV for each frequency band. The values are scaled to the $[0, 1]$ interval for better visualization. Conceptually, as shown in (\ref{eqn: element-wise feature propagation}), the diagonal values in $\mathbf{A}$ represents the contribution of each channel in computing the final EEG representation. It is clear from \ref{fig: activation maps} that there is strong activation on the pre-frontal, parietal and occipital regions for all frequency bands, indicating that these regions may be strongly related to the emotion processing in the brain. Our finding is consistent with existing studies, which observed that asymmetrical frontal and parietal EEG activity may reflect changes on both valence and arousal \cite{schmidt2001frontal, lin2010eeg}. The synchronization between frontal and occipital regions has also been reported to be related to positive emotions \cite{costa2006eeg}. In addition, there is strong activation on the temporal regions for beta and gamma bands, which is consistent with \cite{zheng2015investigating}. The symmetry pattern on the activation maps of channels also indicates that the asymmetry in EEG activity between the left and right hemispheres is critical for emotion recognition.

We identify important inter-channel relations for emotion recognition. Fig. \ref{fig: electrode connectivity} shows the top 10 connections between channels having the largest edge weights in our adjacency matrix $\mathbf{A}$. 
Note that all global connections remain among the strongest connections after $\mathbf{A}$ is learned, demonstrating again that global inter-channel relations are essential for emotion recognition. It is clear from Fig. \ref{fig: electrode connectivity}(b) that the connection between the channel pair (FP1, AF3) is the strongest, followed by (F6, F8), (FP2, AF4) and (PO8, CB2), indicating that local inter-channel relations in the frontal region may be important for emotion recognition.

\section{Conclusion}
In this paper, we propose a regularized graph neural network for EEG-based emotion recognition. Our model is inspired by neuroscience theories on human brain organization and captures both local and global inter-channel relations in EEG signals. In addition, we propose two regularizers, namely NodeDAT and EmotionDL, to improve the robustness of our model against cross-subject EEG variations and noisy labels, respectively. Extensive experiments on two public datasets demonstrate the superior performance of our model than several competitive baselines and the state-of-the-art BiHDM in most experimental settings. 
Our model analysis shows that our proposed biologically inspired adjacency matrix and two regularizers contribute consistent and significant gain to the performance of our model. Investigations on the brain regions reveal that pre-frontal, parietal and occipital regions may be the most informative regions for emotion recognition. In addition, global inter-channel relations between the left and right hemispheres are important, and local inter-channel relations between (FP1, AF3), (F6, F8) and (FP2, AF4) may also provide useful information. 

In the future, we plan to explore: 1) training a more discriminative domain classifier, e.g., by using more advanced classifiers or applying more sophisticated techniques to handle imbalanced samples between training and test sets, to help our model learn more domain-invariant EEG representations; 2)~applying our model to EEG signals that have a smaller number of channels. A simpler version of our model and more advanced regularizations may be necessary to avoid over-smoothing on these small graphs. In addition, data processing techniques that can improve the spatial resolution of EEG signals, e.g., spatial filtering, may be worth exploring.

\ifCLASSOPTIONcompsoc
  \section*{Acknowledgments}
\else
  \section*{Acknowledgment}
\fi
This research is supported by Alibaba Group through Alibaba Innovative Research Program, Alibaba-NTU Singapore Joint Research Institute (Alibaba-NTU-AIR2019B1), Singapore Ministry of Health under its National Innovation Challenge on Active and Confident Ageing (MOH/NIC/COG04/2017; MOH/NIC/HAIG03/2017), the National Research Foundation, Singapore under its NRF Investigatorship Programme (NRF-NRFI05-2019-0002) and under its AI Singapore Programme (AISG Award No: AISG-GC-2019-003). Any opinions, findings and conclusions or recommendations expressed in this material are those of the authors and do not reflect the views of National Research Foundation, Singapore.

\ifCLASSOPTIONcaptionsoff
  \newpage
\fi

\bibliographystyle{IEEEtran}
\bibliography{TAC2019}
\vskip -2\baselineskip plus -1fil
\begin{IEEEbiography}
[{\includegraphics[width=1in,height=1.25in,clip,keepaspectratio]{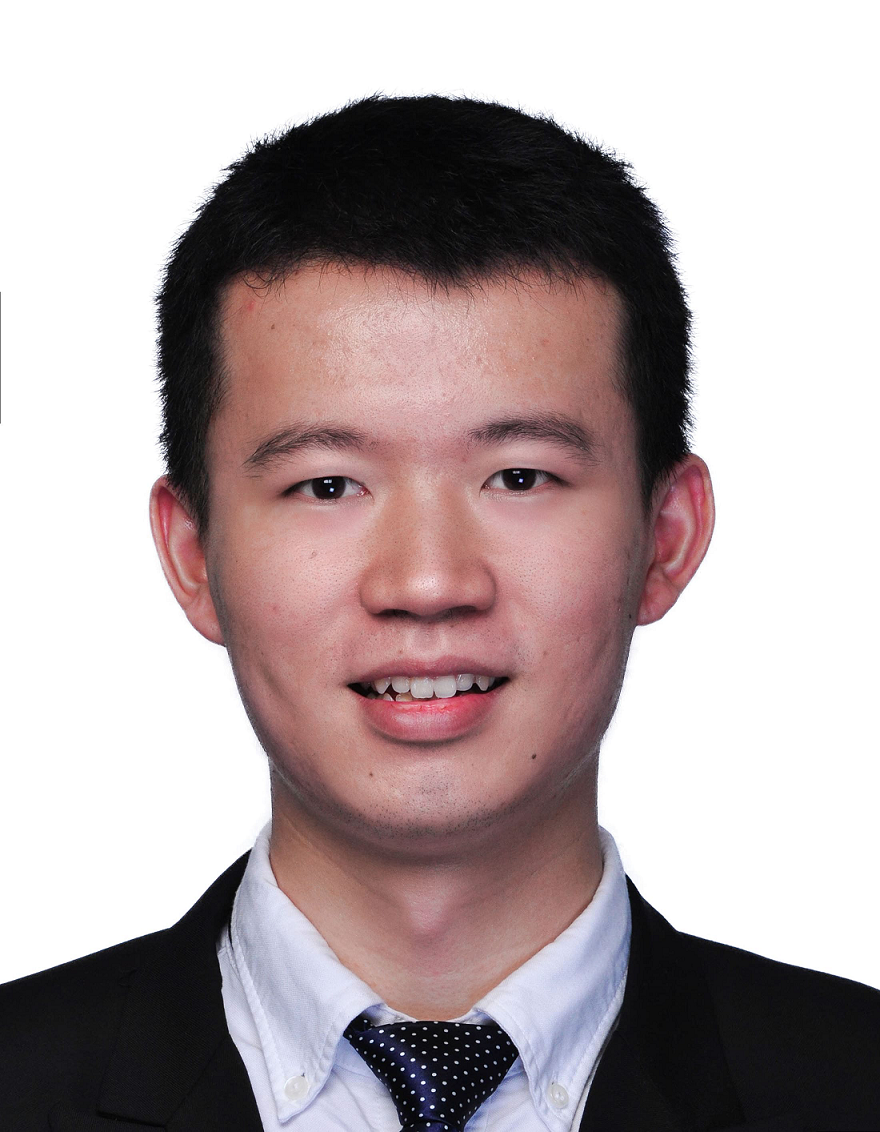}}]{Peixiang Zhong} received the B.Eng. degree in Electrical and Electronic Engineering from Nanyang Technological University, Singapore in 2016. He is currently a PhD candidate in the School of Computer Science and Engineering, Nanyang Technological University, Singapore. His research interests include affective computing, natural language processing and machine learning, etc.
\end{IEEEbiography}
\vskip -2\baselineskip plus -1fil
\begin{IEEEbiography}
[{\includegraphics[width=1in,height=1.25in,clip,keepaspectratio]{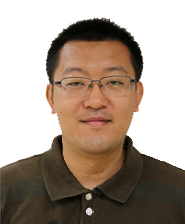}}]{Di Wang} received the B.Eng. degree in Computer Engineering and the Ph.D. degree in Computer Science from Nanyang Technological University, Singapore, in 2003 and 2014, respectively. He is currently working as a Senior Research Fellow and the Research Manager in the Joint NTU-UBC Research Centre of Excellence in Active Living for the Elderly (LILY), Nanyang Technological University, Singapore. His research interests include computational intelligence, decision support systems, computational neuroscience, autonomous agents, affective computing, ubiquitous computing, etc.
\end{IEEEbiography}
\vskip -2\baselineskip plus -1fil
\begin{IEEEbiography}
[{\includegraphics[width=1in,height=1.25in,clip,keepaspectratio]{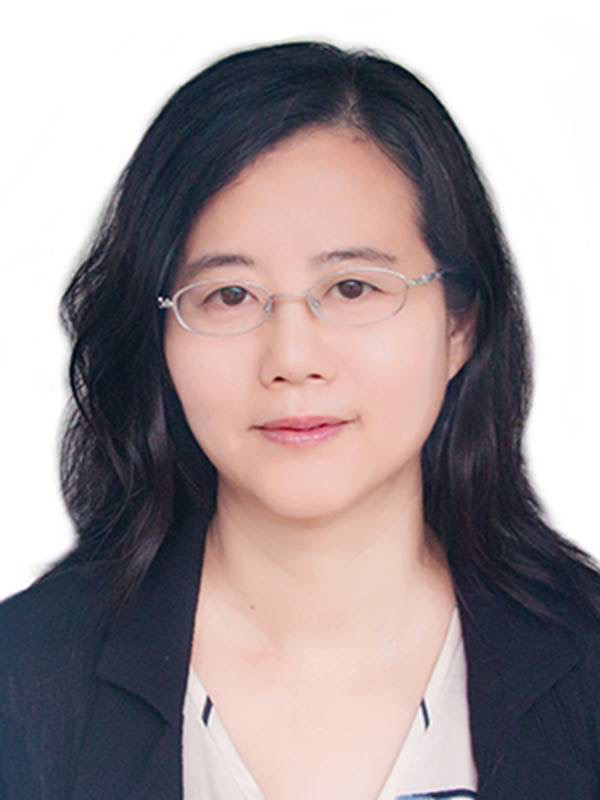}}]{Chunyan Miao} received the B.S. degree from Shandong University, Jinan, China, in 1988, and the M.S. and Ph.D. degrees from Nanyang Technological University (NTU), Singapore, in 1998 and 2003, respectively. She is currently a Professor and the Chair of the School of Computer Science and Engineering, NTU, the Director of the Joint NTU-UBC Research Centre of Excellence in Active Living for the Elderly, and the Director of the Alibaba-NTU Singapore Joint Research Institute. Her current research interests focus on humanized artificial intelligence, which includes infusing intelligent agents into interactive new media (virtual, mixed, mobile, and pervasive media) to create novel experiences and dimensions in game design, interactive narrative, and other real world agent systems.
\end{IEEEbiography}
\end{document}